\documentclass[10pt]{article}

\usepackage[preprint]{tmlr}
\usepackage{xcolor}
\definecolor{linkblue}{RGB}{0,55,100}

\usepackage{latexsym}
\usepackage[T1]{fontenc}
\usepackage[utf8]{inputenc}
\usepackage{microtype}
\usepackage{graphicx}
\usepackage{booktabs}
\usepackage{amsmath}
\usepackage{fvextra}
\usepackage[most]{tcolorbox}
\definecolor{promptblue}{RGB}{24,42,170}
\newtcolorbox{promptbox}[1]{enhanced,breakable,
  colback=promptblue!3!white,colframe=promptblue,colbacktitle=promptblue,
  coltitle=white,fonttitle=\bfseries,title={#1},
  title after break={#1\ (continued)},
  boxrule=0.5pt,arc=2pt,left=8pt,right=8pt,top=6pt,bottom=6pt,
  toptitle=4pt,bottomtitle=4pt,before skip=10pt,after skip=10pt}
\usepackage{flafter}
\usepackage{placeins}
\usepackage{needspace}
\usepackage{float}
\usepackage{array}
\usepackage{hyperref}
\usepackage{url}
\hypersetup{colorlinks=true,linkcolor=linkblue,citecolor=linkblue,urlcolor=linkblue}

\title{From Outcomes to Components: A Counterfactual\\Evaluation Framework for LLM-Proposed Repairs\thanks{GPT assisted with drafting and revising the manuscript. The authors verified the data, analyses, and final text and take responsibility for the work.}}
\author{\name Hengyu Ye \email yehy2025@shanghaitech.edu.cn \\
\addr ShanghaiTech University
\AND
\name Songhua Hu\thanks{\protect\hypertarget{Hfootnote.1}{}Corresponding author.} \email Songhua.Hu@cityu.edu.hk \\
\addr City University of Hong Kong}

\begin{document}
\maketitle

\begin{abstract}

A repair action proposed by a large language model (LLM) specifies which records to change and how to change their values. A failed action may still contain a useful choice of records or a useful update, depending on how the two are paired. We introduce a counterfactual evaluation framework that separates selected records (targets) from the applied change (an update). Four executions combine proposed and reference targets and updates to measure the effects of correcting or retaining each component and their interaction. We evaluate the framework on two tasks: traffic assignment, where affected demand must be identified from road counts, and database repair, where affected rows must be identified from aggregate reports. An injected fault supplies a reference action that reverses it, and the recovery rate measures an action's loss reduction relative to that reference. For sparse-target actions in the traffic experiments, retaining the proposed update on reference targets achieves 79.4\% of the reference loss reduction, compared with 43.7\% from tuning update strength on the proposed targets. In the database experiments, update replacement worsens 17 of 108 actions, all selecting unaffected records, with sharply different frequencies across models. The value of a proposed update depends on the targets it is paired with. Missed targets hide useful updates, while excess targets can make a milder update better than the reference correction. In a component-reuse study, testing new target--update pairings from two proposals raises mean held-out recovery rate to 0.616, compared with 0.580 when selecting between the original proposals.

\end{abstract}

\section{Introduction}\label{1-introduction}

Large language models (LLMs) use tools to inspect and modify external systems \citep{lu2025toolsandbox,yao2024taubench}. Repair tasks require them to turn observed discrepancies into changes that restore the system's behavior \citep{jha2025itbench}. For repairs that act on selected records, \textbf{target selection} identifies which records to change, and \textbf{update selection} specifies the change to their values. An unsuccessful repair may contain a useful update applied to the wrong targets. Improving the next repair requires deciding what to change and what to retain.

Existing evaluations measure task outcomes and locate failures \citep{lu2025toolsandbox,zhang2025whowhen}. An outcome score reports whether a repair worked. Failure attribution locates the step where it failed. Both treat the action as a unit. Studies of reasoning separate finding an error from correcting it \citep{tyen2024localization}, while counterfactual evaluations assess targeted interventions \citep{ma2026dover,bonagiri2026causalflow}. For a model-proposed repair, we ask: \textbf{How does the value of a proposed target set or update depend on the component it is paired with?}

We examine traffic assignment, where demand must be inferred from road counts, and database repair, where affected rows must be inferred from aggregate reports. Target selection errors can affect the value of an update in different ways. Missed targets can obscure a useful update by leaving affected records unchanged. Conversely, when a proposal selects too many records, a mild update may limit damage to unaffected records. Replacing it with the inverse of the fault can then repair affected records more fully while damaging unaffected records. The effect of correcting one decision depends on its partner.

A discrepancy can involve many records, and changing one record can affect several observations. Models must infer both where to intervene and how much to change from this indirect evidence. Road counts combine trips from many origin--destination (OD) pairs, while database reports combine rows with overlapping attributes.

\begin{figure}[ht]
\centering
\includegraphics[width=\textwidth]{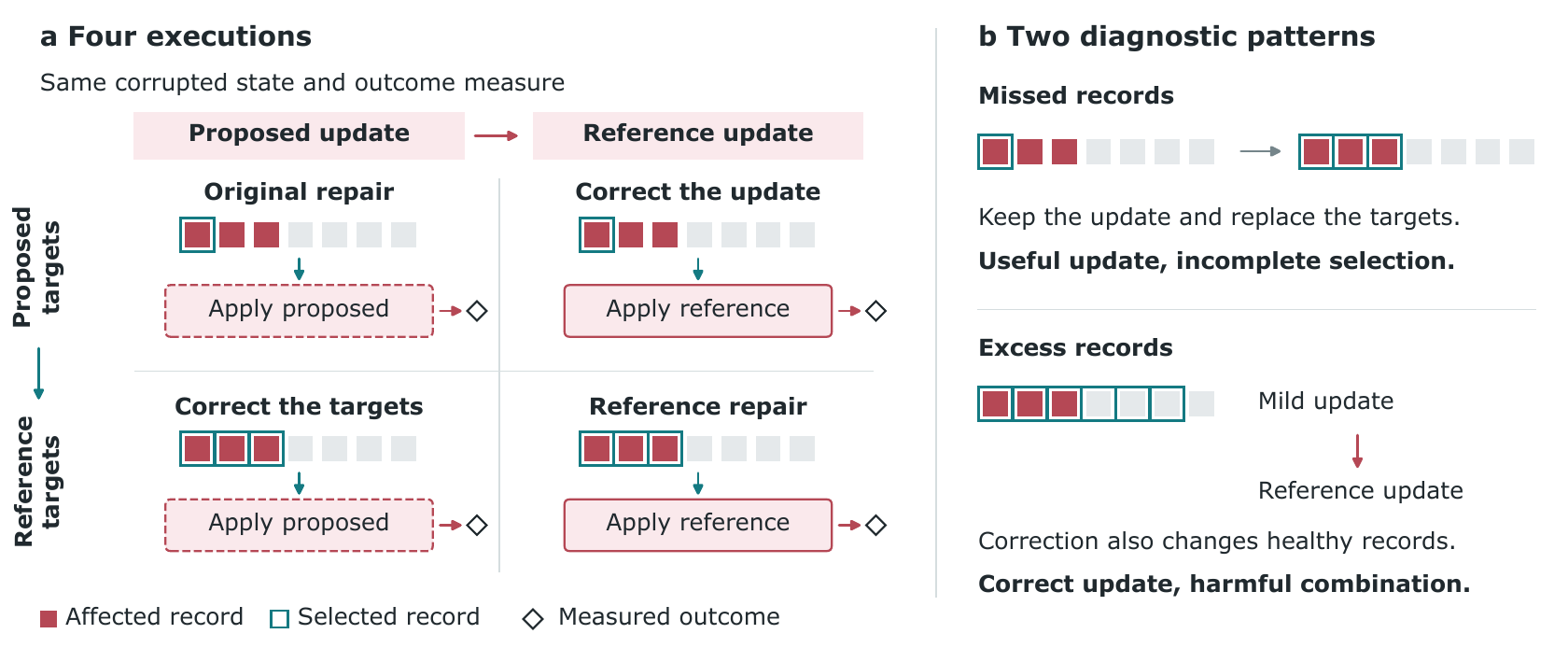}
\caption{\textbf{From a repair outcome to a diagnosis of its decisions.} A repair action supplies selected records and a numerical update. Four executions combine proposed and reference components, changing the update horizontally and the targets vertically. The illustrated selection errors lead to different diagnoses. Missed records can hide a useful update, while excess records can make a stronger correction harmful. Record diagrams are schematic. Reference components enter the evaluation after the model submits its action.}
\label{fig:grounding-overview}
\end{figure}

We propose a framework for evaluating these decisions within an already generated repair (Figure~\ref{fig:grounding-overview}). \textbf{Component replacement} exchanges either the proposed targets or the proposed update for its reference counterpart, leaving the other unchanged. Four executions from the same initial state measure the proposed action, each replacement, and their combination. The comparison tests both whether a proposed component is useful with the reference partner and whether correcting it helps with the proposed partner. We instantiate the framework with known faults, whose inverse actions provide the reference components. Two proposals for the same fault contribute each other's targets and updates, and executed outcomes rank the candidates they form.

The study makes three contributions:
\begin{itemize}
\item \textbf{A counterfactual evaluation framework} that distinguishes retained component value from correction effects through executable target--update replacements.
\item \textbf{Cross-domain diagnostic evidence} that missed targets hide useful updates, while excess targets can make update correction harmful. Fixed-target multiplier search measures what update tuning can recover, and repeated database trials measure how often update correction causes harm.
\item \textbf{A component-reuse study} that exchanges decisions between proposals and selects repairs through execution preview. Held-out observations measure when the additional candidates help, how gains vary across generations, and the execution cost of obtaining them.
\end{itemize}

\Needspace{8\baselineskip}
\section{Related Work}\label{11-relation-to-existing-evaluations}

\textbf{Decisions within failed responses.} The compositionality gap separates answering constituent questions from combining their answers \citep{press2023compositionality}. \citet{tyen2024localization} distinguish locating reasoning errors from correcting them through backtracking from a supplied correct location. Thought-ICS structures reasoning into discrete thoughts for self-localization and correction \citep{samanta2026structure}. STOCKTAKE separates state estimation from control outcomes \citep{deb2026stocktake}.

\textbf{Outcome and counterfactual evaluation.} ToolSandbox and \(\tau\)-bench evaluate changes to persistent state, while AgentBoard measures progress over an episode \citep{lu2025toolsandbox,yao2024taubench,ma2024agentboard}. The Who\&When dataset evaluates failure attribution by agent and trajectory step \citep{zhang2025whowhen}. DoVer tests targeted changes, CausalFlow evaluates local repairs through execution or outcome prediction, and joint replay identifies combinations of supplied clean-message replacements that restore success \citep{ma2026dover,bonagiri2026causalflow,li2026jointreplay}. ORACLE-SWE measures the effects of supplying correct information, including edit locations and tests, separately and in combination \citep{li2026oracleswe}. ShapleyFlow attributes gains and interactions to model upgrades across workflow modules \citep{yang2026shapleyflow}.

\textbf{Structured repairs.} QFix repairs constants in selection predicates and update expressions in query histories using reported record errors \citep{wang2017qfix}. Castle generates direct and cascading column updates from database schemas \citep{su2025castle}. For model-generated actions, ReflecTool-Bench evaluates tool-error detection and correction, and Action Boundary Blindness examines action scope and completeness \citep{liu2026reflective,wang2026boundary}.

\textbf{Candidate construction and selection.} Execution-guided decoding filters SQL candidates using execution errors and empty results \citep{wang2018execution}, and tree search uses executed web actions with a model-based value function \citep{koh2025search}. Simulation feedback supports physical-parameter inference in SimLM \citep{memery2024simlm} and physical-law discovery in PhysGym \citep{chen2025physgym}. Mind Evolution combines evaluator-guided selection with model-generated recombination and refinement \citep{lee2025mindevolution}.

Prior work identifies failing agent steps or tests interventions on information signals and event sets \citep{ma2026dover,bonagiri2026causalflow,li2026oracleswe,li2026jointreplay}. Database repair methods also modify selection predicates and update expressions \citep{wang2017qfix}. They leave unresolved which decisions in a failed repair remain useful, and whether correcting one decision can make the action worse. We address this question by independently replacing targets and updates and comparing their executed outcomes. We then show that crossing two proposals for the same fault yields candidate repairs that can outperform both originals.

\section{Component-Replacement Framework}\label{2-action-grounding-task-and-controlled-testbed}

\subsection{Problem setup}
Each task starts from a healthy traffic or database state. We inject a fault that changes selected input records and produces a corrupted state $x_0$. The model observes aggregate discrepancies between the corrupted and healthy outcomes, then proposes a repair. We call the records chosen for editing \textbf{targets}: OD pairs or links in traffic and rows in the database. For numerical repairs, the injected fault specifies a reference action that reverses its changes. This reference is available to the evaluator after the proposal and defines the component comparisons.

\subsection{Component interventions on a proposed repair}\label{21-from-diagnostic-evidence-to-a-structured-intervention}

An action $\hat a=(\hat S,\hat u)$ applies an update $\hat u$ to targets $\hat S$. Evaluation pairs it with the reference action $a^\star=(S^\star,u^\star)$.

Let $J_0$ be the loss before repair, $J_p$ the loss after the proposed action, and $J^\star$ the loss after the reference action. The \textbf{recovery rate} is
\begin{equation}
  \label{eq:recovery}
  R = \frac{J_0-J_p}{J_0-J^\star}.
\end{equation}
A recovery rate of zero means no improvement, one matches the reference repair, and $-1$ adds as much loss as the reference removes. Values can fall below $-1$ when damage is larger, or exceed one when the proposed repair achieves a lower loss than the reference action.

The framework replays four actions from $x_0$:
\begin{equation}
  (\hat S,\hat u),\quad(\hat S,u^\star),\quad(S^\star,\hat u),\quad(S^\star,u^\star).
  \end{equation}
The first action is the proposed action, the next two replace one component at a time, and the last is the reference action. Their recovery rates are $R_{00}$, $R_{01}$, $R_{10}$, and $R_{11}$, respectively. The first subscript gives the target source and the second the update source, with 0 for proposed and 1 for reference. By normalization, $R_{11}=1$. Each single-component replacement preserves the other component exactly. Proposed and reference updates operate on the same type of editable record. The loss and observation panel remain fixed across executions. Holding the other proposed component fixed, the direct effects of target and update replacement are $R_{10}-R_{00}$ and $R_{01}-R_{00}$, respectively. Their interaction is
\begin{equation}
I=R_{11}-R_{10}-R_{01}+R_{00}.
\label{eq:component-interaction}
\end{equation}
Positive $I$ means that replacing one component increases the gain from replacing the other. Negative $I$ means that it decreases that gain. Appendix~\ref{a-component-search} gives the symmetric component allocation used for aggregate comparisons.

\textbf{Reading the comparison.} Target accuracy measures agreement with $S^\star$. $R_{10}$ measures recovery with the proposed update on reference targets, and $R_{10}-R_{00}$ isolates the effect of replacing the targets. $R_{01}-R_{00}$ isolates the effect of correcting the update. These comparisons separate retained component value from correction effects. An update that works on reference targets can therefore be useful even when its original action fails. We call an update \textbf{compensating} when it limits harm from excess targets, with a negative $R_{01}-R_{00}$ as its observed signature: replacing that update worsens the repair.

\textbf{Instantiation requirements.} The target and update components must be independently executable, all four actions must start from the same state, and their outcomes must use the same loss. Reference components specify the comparison being made and can be chosen for a particular repair task.

\section{Evaluation Settings and Metrics}\label{evaluation-settings}
\subsection{Traffic repair}\label{22-why-traffic-assignment-provides-a-useful-testbed}
Traffic assignment couples OD pairs through shared routes.

The traffic environment has two input tables. Each row of the \textbf{demand table} stores the demand for one OD pair: the number of trips from an origin to a destination. A \textbf{capacity table} stores the capacity of each directed road segment, or link. Traffic assignment routes trips through the network and returns link flows, the number of trips traversing each road segment. Link travel times rise with assigned flow, so changing OD demand also changes route choices and the flows generated by other OD pairs. The agent selects targets and multiplies their current values by a chosen factor.

The agent must trace discrepancies in road counts back to the OD pairs that need repair. The experiments use the Sioux Falls network \citep{leblanc1975traffic}. Experiments with fixed diagnostic evidence, action feedback, and candidate construction replace identifiers with anonymous aliases and perturb healthy demand values. Appendix~\ref{a7-tool-catalogue-and-benchmark-difficulty-details} describes the simulator and construction.

Faults decrease or increase trip counts, reduce road capacity, or distort sensor readings. A healthy control preserves the original state. A demand or capacity fault multiplies the affected demand or capacity values by $m$, so its reference update is $1/m$. Sensor repair identifies biased observations for exclusion. Agents inspect demand summaries, road capacities, and residuals between current simulated flows and observed road counts. For demand and capacity repairs, agents receive road counts from the healthy assignment and summaries of the current inputs. The component replacement analysis covers the demand-scaling actions in traffic and the numerical repairs in the database. Sensor bias is evaluated through fault recognition and target accuracy, while the healthy control is evaluated through fault recognition alone.

\subsection{Database repair}\label{database-setting}

In database repair, the agent identifies which rows cause discrepancies in aggregate reports and chooses how to update them. The environment uses DuckDB \citep{raasveldt2019duckdb} and 600,572 transaction records generated from the Transaction Processing Performance Council's decision-support benchmark (TPC-H) \citep{tpc2022tpch}. Each record contains an amount, an adjustment rate, and categorical attributes. Reports aggregate records by these attributes. The agent receives aggregate discrepancies relative to reports from the healthy database. Actions specify a row-selection predicate and a numerical update. A predicate combines categorical conditions, and an update multiplies or adds to one field on the selected rows. Healthy values are perturbed before fault injection, and identifiers are replaced with anonymous aliases. Appendix~\ref{a-database} gives the construction, action grammar, and reporting workload.

\subsection{Evaluation metrics}\label{3-measuring-recognition-and-repair}

We evaluate repairs through three measures: fault recognition, target accuracy, and recovery rate after execution. Fault recognition and target accuracy score the action before execution. The recovery rate scores its effect. Component replacement then attributes that effect to each decision (Section~\ref{21-from-diagnostic-evidence-to-a-structured-intervention}).

\textbf{Fault recognition.} In the traffic study, fault recognition requires an executable action that correctly identifies the fault class, its direction, and the corresponding repair operation. Before execution, the validator checks that the action uses the expected JSON keys, references existing records or sensors, and supplies numerical values within the allowed ranges.

\textbf{Target accuracy.} Precision is the fraction of selected records that belong to the injected target set. Recall is the fraction of affected records selected, and target F1 is their harmonic mean. For demand repairs, positive-demand recall is the fraction of affected OD pairs with nonzero demand that are selected.

\textbf{Outcome losses.} Traffic loss uses the GEH statistic, which scales the difference between simulated and reference counts by their combined volume \citep{transportscotland2008geh}. For a set of observed links $P$, let $q_\ell$ and $y_\ell$ denote simulated and reference counts on link $\ell$. The loss averages squared GEH:
\begin{equation}
J_{\mathrm{traffic}}=\frac{1}{|P|}\sum_{\ell\in P}
\frac{2(q_\ell-y_\ell)^2}{q_\ell+y_\ell+\epsilon_{\mathrm{traffic}}}.
\label{eq:traffic-loss}
\end{equation}
Database reports aggregate amounts, adjustment rates, and adjusted amounts. An adjusted amount is a record's amount multiplied by one minus its adjustment rate. Let $Q$ index the aggregate values used for scoring, with $z_k$ and $y_k$ denoting the current and healthy values of aggregate $k$. Loss averages squared relative errors:
\begin{equation}
J_{\mathrm{database}}=\frac{1}{|Q|}\sum_{k\in Q}
\frac{(z_k-y_k)^2}{(|y_k|+\epsilon_{\mathrm{database}})^2}.
\end{equation}
Both constants provide numerical protection against zero denominators. Equation~\ref{eq:recovery} normalizes improvement by the reference action in each domain. A traffic recovery rate is \textbf{near-complete} when an action achieves at least 90\% of the reference action's loss reduction ($R\geq0.9$), and \textbf{harmful} when it increases loss ($R<0$). Sensor repair uses fault recognition and target accuracy. Appendices~\ref{a7-tool-catalogue-and-benchmark-difficulty-details} and~\ref{a-database} specify the constants, simulator, and reporting workload.

\Needspace{12\baselineskip}
\noindent\begin{minipage}[t]{\linewidth}
\section{Experimental Design}\label{4-experimental-design}

The experiments evaluate four questions about the framework:
\begin{enumerate}
\setlength{\itemsep}{2pt}
\setlength{\parskip}{0pt}
\item \textbf{Diagnostic need.} What do fault recognition and target accuracy leave unresolved about repair outcomes?
\item \textbf{Retained value.} Can a failed action contain a useful update, and can tuning that update compensate for the proposed targets?
\item \textbf{Partner dependence.} When does correcting one component harm the action, and how does this depend on selection errors?
\item \textbf{Component reuse.} Can crossing decisions from two proposals improve held-out repair relative to selecting one proposal, and what is the execution cost?
\end{enumerate}
\end{minipage}
\par

\textbf{Models.} To compare fault recognition with recovery outcomes, the traffic evaluation covers 11 models on seven conditions. They are GPT-6 Astra (Astra), GPT-5.6 Sol (Sol), and GPT-5.6 Terra (Terra), Claude Opus 4.8 (Opus) and Claude Sonnet 4.6 (Sonnet), Gemini 3.8 Flash (Gemini), Kimi K3 (K3), DeepSeek-V4.1-Flash-1017 (DeepSeek-Flash), GLM-5.3 and GLM-5.3-Flash (GLM-Flash), and MiMo-V2.5-Pro (MiMo-Pro) \citep{openai2026models,anthropic2026cards,google2026flash,kimi2026k3,deepseek2026flash,zai2026models,xiaomi2026mimo}. Appendix~\ref{a2-prompts-and-endpoints} records endpoint identifiers, dates, decoding settings, and provider routes.

\textbf{Fault recognition and difficulty.} All 11 models inspect evidence through a common tool catalogue and return one final action for each of seven conditions: four demand faults, a capacity reduction, sensor bias, and a healthy control. A difficulty study then sweeps fault magnitude and the spatial distribution of affected records over six seven-level series and a healthy control for K3, DeepSeek-Flash, GLM-5.3, GLM-Flash, and MiMo-Pro. The sweep varies fault magnitude and spatial distribution while preserving fault-class visibility, giving the graded range of target-selection quality and repair outcomes that the component analysis uses. Its demand episodes supply the actions analyzed in Section~\ref{52-which-action-component-limits-repair}. Appendix~\ref{difficulty-profile-results} reports the profiles, and Appendix~\ref{difficulty-grids-and-certification} gives the instance construction and parameter values.

\textbf{Component analysis across domains.} We apply the four-action comparison in Section~\ref{21-from-diagnostic-evidence-to-a-structured-intervention} to traffic and database repairs.

In the traffic experiments, the difficulty study's 140 demand episodes produce 117 demand-scaling actions. Three have empty targets, leaving 114 actions for component analysis.

To measure the effect of update tuning on missed targets, we analyze 54 nonempty demand-scaling actions from eight fault cases. They cover few affected OD pairs (sparse-target actions) and comprise 28 distinct state--target combinations, because several actions start from the same state and select the same targets. For each selected target set, fixed-target search varies the multiplier over its legal range and compares the best tested recovery rate with target replacement. This measures how much numerical tuning compensates for incomplete selection. Appendices~\ref{a-component-search} and~\ref{a7-tool-catalogue-and-benchmark-difficulty-details} specify eligibility, sampling sources, the search procedure, and the solver settings.

The database study examines the same target--update interactions in repairs to rows contributing to aggregate reports. Six faults alter amounts or adjustment rates within intersections of categorical groups. K3, GLM-5.3, and DeepSeek-Flash each repair every fault under two evidence conditions in three generation rounds, yielding 108 actions. Each round requests a fresh action for every model--fault--evidence combination using the same inputs and generation settings. Direct evidence includes a report aggregated over the affected combination of categories. Time-grouped evidence replaces that report with one grouped by time, so the affected category intersection is no longer isolated in the displayed report. Repeated generations and evidence conditions share the same six underlying faults. Appendix~\ref{a-database} specifies the report construction and gives all three rounds of component outcomes.

\textbf{Component reuse with execution preview.} To examine alternative combinations of existing proposals, K3, DeepSeek-Flash, and GLM-5.3 each generate two traffic repairs: one lists individual OD pairs and the other selects origin groups, which expand to all positive-demand OD pairs from those origins. Execution preview ranks the originals and candidates constructed by exchanging or interpolating their components. The experiment uses four demand faults and three independent generations per condition. The 76 observed links form fixed panels A and B of 38 sensors each. In one panel direction, A is the selection panel: it supplies generation evidence and ranks executed candidates. B is the evaluation panel and measures the selected repair. The reverse direction exchanges their roles and uses fresh proposals because the generation evidence changes. Section~\ref{53-does-assistance-change-the-limiting-component} reports the results, and Appendix~\ref{a3b-target-representation} gives the protocol. Table~\ref{tab:analysis-sets} lists the actions entering each analysis.

\begin{table}[!htbp]
\centering
\caption{\textbf{Analysis sets and eligibility.} Counts refer to actions unless stated otherwise.}
\label{tab:analysis-sets}
\begin{tabular}{p{0.24\textwidth}p{0.69\textwidth}}
\toprule
Analysis & Included observations \\
\midrule
Traffic survey & 77 episodes from 11 models and seven conditions. Five numerical-repair conditions yield 55 actions. Sensor bias and the healthy control contribute fault recognition outcomes. \\
Traffic components & 140 demand episodes yield 117 demand-scaling actions. Three empty target sets leave 114 for component analysis. Fixed-target search uses 11 fixed-evidence and 43 feedback actions, covering 28 distinct state--target combinations. \\
Database components & 108 actions from three models, six faults, two evidence conditions, and three generations. \\
Traffic preview & 72 directional evaluations from three models, four faults, three generations, and two panel directions. Averaging directions yields 36 paired results. \\
\bottomrule
\end{tabular}
\end{table}

\section{Results}\label{5-results}

\subsection{Fault recognition and repair outcomes}\label{51-does-recognition-lead-to-effective-repair}

\begin{figure}[!htb]
\centering
\includegraphics[width=\textwidth]{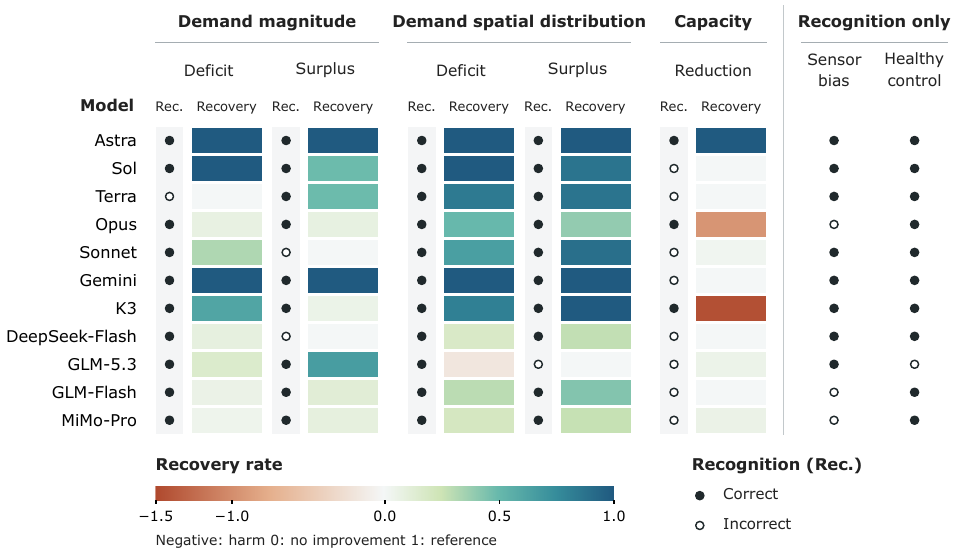}
\caption{\textbf{Fault recognition coexists with widely varying recovery rates across 11 models.} Each row shows one model across seven conditions. Filled and open circles denote correct and incorrect fault identification on 77 interactive episodes. Adjacent colored tiles show recovery rates for all 55 numerical-repair episodes across fault recognition outcomes. Colors saturate outside $[-1.5,1]$. Sensor bias and the healthy control show fault recognition only.}
\label{fig:model-survey-results}
\end{figure}

\textbf{Recognizing the fault is easier than repairing it.} Across 11 models, fault recognition is much more frequent than a near-complete recovery rate (Figure~\ref{fig:model-survey-results}). The five demand and capacity conditions contribute 55 of the 77 episodes to the numerical-repair comparison. Of these 55 actions, 43 (78.2\%) identify the fault, its direction, and the repair operation, and 12 (21.8\%) achieve a near-complete recovery rate. On the same demand deficit, Astra and K3 select the same 161 OD pairs. Their multipliers of 1.620 and 2.000 yield recovery rates of 1.000 and 0.612, respectively. The larger multiplier overcorrects demand on those pairs. Changing demand also changes equilibrium route choices, so the same targets can produce different road flows and repair outcomes.

\textbf{Aggregate residuals point to road endpoints.} Action traces show the distinction between recognizing a discrepancy and selecting its source. In the fixed-evidence study, five models receive the same precomputed diagnostic information for each condition. In the broad-deficit condition, four of them select OD pairs that match road endpoints in the residual report. Those road counts aggregate trips from many OD pairs. Models must infer the affected pairs from the available identifiers and aggregate evidence (Appendix~\ref{a1-fixed-evidence-evaluation}). Identical targets with different multipliers reach different recovery rates, and correct fault recognition with wrong targets still fails.

Fault recognition stays constant while repair outcomes change. K3 recognizes every demand-deficit condition as the affected set shifts from seven contiguous origin zones to ten dispersed OD pairs at a fixed fault multiplier. Target F1 falls from 0.979 to 0.621, and the recovery rate falls from 1.000 to $-1.276$. Figure~\ref{fig:difficulty-ladder-results} extends this comparison across five models and six difficulty series.

\FloatBarrier
\subsection{Failed repairs retain useful updates}\label{52-which-action-component-limits-repair}

\textbf{Target and update corrections interact.} In the component analysis of 114 demand-scaling actions from the difficulty study, target and update corrections have positive interaction ($I>0$) in 76 actions. Correcting one component then increases the gain from correcting the other. In six actions, either single-component replacement lowers the recovery rate by more than 0.01, yet replacing both improves it. The errors in the original target and update decisions partly compensate for each other. We average each replacement effect over both partner settings. Target correction then contributes more when target F1 is low, and update correction contributes more when targets are nearly correct (Appendix~\ref{a-component-search}).

\begin{figure}[!htbp]
\centering
\includegraphics[width=\textwidth]{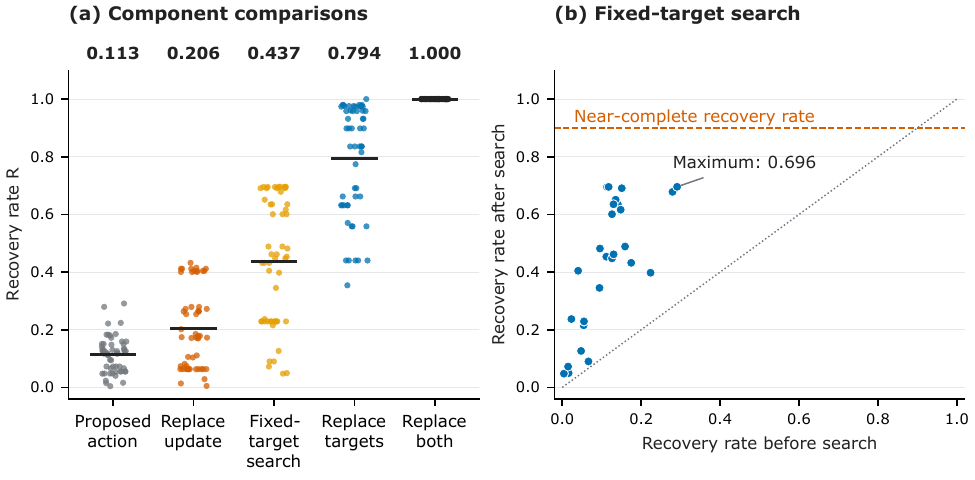}
\caption{\textbf{Component replacement and fixed-target search on sparse targets.} Left: recovery rates across 54 actions. Horizontal bars and numbers above the groups show means. Right: recovery rates before and after search for 28 distinct state--target combinations. The before-search rate uses the first recorded action for each combination. Each search evaluates 28 multipliers while preserving the selected targets.}
\label{fig:complete-component-replacement}
\end{figure}

\textbf{Retaining the proposed update does more than tuning it.} Fixed-target search tests whether a better multiplier can rescue a sparse-target action without adding missed OD pairs. It tests 28 legal multipliers per distinct state--target combination: 25 logarithmically spaced values and three local refinements. Selected OD pairs cover 9.0\% of affected pairs with positive demand on average. Search raises the mean recovery rate from 0.113 to 0.437. Replacing the targets and preserving the model's multiplier reaches 0.794 (Figure~\ref{fig:complete-component-replacement}). Target replacement outperforms fixed-target search in 43 of 54 actions. The remaining 11 actions all come from the same two-origin demand-deficit state, and the highest recovery rate reached by any fixed-target search is 0.696.

\FloatBarrier
\subsection{Excess targets can turn update correction into collateral damage}\label{database-results}

In contrast to the sparse-target failures in traffic, excess targets in database repair expose unaffected rows to the update. On excess targets, a milder proposed update can limit damage to unaffected rows. Applying the reference update to the same targets can restore affected rows while damaging unaffected rows. Across 108 database actions, 33 have inaccurate targets. Target replacement improves the recovery rate by more than 0.01 in 27 of them. Replacing only the update worsens 17 actions by more than 0.01. All 17 select unaffected rows, and 16 also cover every affected row.

This pattern differs across models. DeepSeek-Flash has excess targets in 21 of 36 actions, and update replacement harms 15 of them, including seven, three, and five actions across the three repetitions. The corresponding counts are four and two for K3, and one and zero for GLM-5.3 (Table~\ref{tab:database-model-summary}). Appendix~\ref{a-database} summarizes the database component results.

\textbf{The available evidence changes target selection.} The two evidence conditions hold the corrupted database and action interface fixed and vary the report grouping. Across 54 model--fault--repetition comparisons, time-grouped evidence lowers target F1 in 19, raises it in four, and leaves it unchanged in 31. The median original-action recovery rate is 1.000 under direct evidence and 0.514 under time-grouped evidence. Evidence grouping changes which records the model selects, and the selected records are the partner of the update in every execution.

\textbf{Harm occurs with both matching and differing operators.} All 17 harmed actions edit the same field as the reference repair, and ten of them use the same operator. In those ten, the reference update corrects the affected rows and also reaches the unaffected rows in the selected set, so recovery falls while the operator stays the same.

Table~\ref{tab:database-components} illustrates these effects with two repairs. Using time-grouped evidence, DeepSeek-Flash selects an entire mode category for a fault confined to one mode--handling intersection. Its proposed update reduces amounts by 7.7\%. The reference update reduces them by 25.4\%, restoring affected rows and causing greater damage to unaffected rows. GLM-5.3 selects exactly the affected rows in another case and proposes a harmful 
numerical update. Update replacement raises the recovery rate to 1.000.

\begin{table}[!ht]
\centering
\caption{\textbf{Database target errors and update-replacement harm.} Each model contributes 36 actions: six faults, two evidence conditions, and three repetitions. Harm means that replacing only the update decreases the recovery rate by more than 0.01. Repetitions share the same faults.}
\label{tab:database-model-summary}
\begin{tabular}{lrrrr}
\toprule
Model & \shortstack{Excess\\targets} & \shortstack{Harm among\\all actions} & \shortstack{Harm with\\excess targets} & \shortstack{Harm counts\\by repetition} \\
\midrule
DeepSeek-Flash & 21/36 & 15/36 & 15/21 & 7, 3, 5 \\
K3 & 4/36 & 2/36 & 2/4 & 1, 0, 1 \\
GLM-5.3 & 1/36 & 0/36 & 0/1 & 0, 0, 0 \\
\bottomrule
\end{tabular}
\end{table}

\Needspace{8\baselineskip}
\subsection{Reusing components through execution preview}\label{53-does-assistance-change-the-limiting-component}

Sections~\ref{51-does-recognition-lead-to-effective-repair}--\ref{database-results} use reference components to identify which decisions inside a failed repair retain value. The final experiment tests whether executed outcomes can rank candidates assembled from more than one proposal.

In the traffic setting, each model supplies two proposals for the same fault: an OD-pair proposal $(S_1,u_1)$ and an origin-grouped proposal $(S_2,u_2)$. \textbf{Component crossing} adds $(S_1,u_2)$ and $(S_2,u_1)$. \textbf{Update interpolation} instead adds $(S_1,\bar u)$ and $(S_2,\bar u)$, where $\bar u=\sqrt{u_1u_2}$ is the midpoint on a logarithmic scale. Each expanded set contains these two candidates and the two originals. Execution preview ranks candidates by selection-panel loss, and the selected repair is measured on the evaluation panel (Table~\ref{tab:preview-main}). Each expanded set requires four candidate evaluations, two more than selection between the originals.

\textbf{Crossed components improve on selection.} The mean recovery rate on the evaluation panels is 0.030 for the OD-pair proposal and 0.506 for the origin-grouped proposal. Selecting between them reaches 0.580. The four-candidate midpoint set reaches 0.605, and the crossed-component set reaches 0.616 (Table~\ref{tab:preview-main}). Crossing gains 0.036 over selection between the originals. It gains 0.011 over midpoint with the same two model generations and four candidate executions. The crossed-set gain over the originals is 0.026, 0.076, and 0.007 in the three repetitions, with a standard deviation of 0.036 across repetition means.

Across the 72 directional evaluations, the crossed set improves on selection between the originals in 16 and ties in 56. The midpoint set improves in 19, ties in 51, and worsens in two.

\begin{table}[ht]
\caption{\textbf{Execution preview.} Mean held-out recovery rate across 36 model--fault--repetition results, each averaged over both panel directions.}
\label{tab:preview-main}
\begin{center}
\begin{tabular}{lrr}
\toprule
Candidate set & Candidates & Mean recovery rate $R$ \\
\midrule
Original OD-pair proposal & 1 & 0.030 \\
Original origin-grouped proposal & 1 & 0.506 \\
Two original proposals & 2 & 0.580 \\
Originals + midpoint updates & 4 & 0.605 \\
Originals + crossed components & 4 & 0.616 \\
\bottomrule
\end{tabular}

\end{center}
\end{table}

\begin{figure}[ht]
\centering
\includegraphics[width=0.92\textwidth]{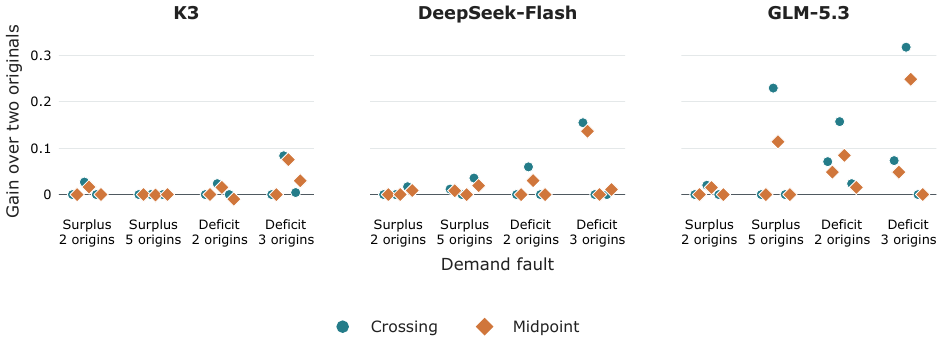}
\caption{\textbf{Component-reuse gains by model and fault.} Three points per model--fault cell show gain over selection between the originals. Each point averages both panel directions. Crossing and midpoint use two proposals and four candidate executions.}
\label{fig:preview-paired-gains}
\end{figure}

Figure~\ref{fig:preview-paired-gains} resolves the aggregate into 36 model--fault--repetition comparisons, each averaged over both panel directions. Mean crossing gains are 0.012 for K3, 0.023 for DeepSeek-Flash, and 0.074 for GLM-5.3. The surplus faults affecting two and five origins yield mean gains of 0.007 and 0.031. The deficit faults affecting two and three origins yield 0.037 and 0.070. The largest paired gain is 0.317 for GLM-5.3 on the three-origin deficit in repetition 2.

In the GLM-5.3 three-origin deficit example, the OD-pair proposal selects five OD pairs with multiplier 1.250. The origin-grouped proposal selects five origins, expanding to 103 positive-demand OD pairs, with multiplier 1.450. Preview selects the OD-pair original over the origin-grouped original on panel A. Crossing the origin-grouped targets with the OD-pair update produces the lowest selection loss and raises held-out recovery rate to 0.680 (Table~\ref{tab:preview-crossing-example}). The other panel direction ties the original selection, yielding the paired gain of 0.317 in Figure~\ref{fig:preview-paired-gains}. The example shows that the origin-grouped target set and OD-pair update work better together even though neither proposal performs well.

\FloatBarrier

\section{Discussion}\label{6-discussion}

The component results reveal two forms of dependence between targets and updates. With missed targets, a useful update has little effect until it reaches the affected records. With excess targets, a mild update can limit damage to unaffected records, so replacing only the update can worsen the repair. The value of either decision therefore depends on the other. Execution preview uses this dependence by testing target--update combinations from two proposals. Its gains concentrate in GLM-5.3 and the three-origin deficit, where the grouped targets and record-level update can work better together than either proposal.

The findings support two uses with different information requirements. When a known fault supplies a reference action, four executions identify which proposed component retains value and how the components interact. When reference components are unavailable for selection, execution preview compares two original proposals with candidates formed by crossing their targets and updates. This uses the same two model generations and four candidate executions, two more executions than selecting between the originals. The extra cost matters when repeated execution is expensive or the selection observations are a weak guide to the final repair outcome. The first use explains a completed repair once the fault is known. The second constructs a better candidate before commitment.

\Needspace{10\baselineskip}
\section{Conclusion}\label{8-conclusion}

Failed LLM repairs can contain useful decisions. In the traffic experiments, retaining the proposed update while replacing sparse targets achieves higher recovery than tuning the update on the original targets. In the database experiments, replacing the proposed update worsens 17 of 108 actions, all of which select unaffected records. The frequency varies across models. The component-replacement framework distinguishes retained value from correction effects by evaluating each decision with different partners. Execution preview further shows that crossing targets and updates from two proposals raises mean held-out recovery from 0.580 to 0.616.

\section*{Limitations}\label{limitations}

\textbf{Settings and actions.} The study uses one static traffic network and one generated relational database with planted faults. Numerical traffic repairs apply one multiplier across the selected records. Database repairs apply one multiplier or additive correction across the selected rows. Fixed-target search evaluates 28 multipliers over a specified range.

\textbf{Sampling.} The traffic model comparisons use one retained action per model and condition. The feedback study also includes revisions. The traffic component analysis contains 114 actions across 28 conditions. The database study contains 108 actions from six faults, three models, two evidence conditions, and three generation rounds. Repetitions measure generation variability on these faults. The preview experiment covers four traffic faults and three models with three repetitions. Its reported range and standard deviation summarize variation across three generations on these four faults. Panel directions and generations are nested within faults, so broader uncertainty remains unestimated.

\textbf{Interfaces and preview.} Grouped selection combines a shorter action representation with the origin structure of the traffic faults. Preview requires repeatable execution and an observable outcome before commitment. Here it compares fixed sets of two or four candidates. The database experiments focus on component effects, while the interface and preview experiments use traffic.

\textbf{Reference-based evaluation.} Component comparisons and recovery rates depend on the reference repair, specified here by fault injection. Alternative valid repairs may yield different component effects. Naturally occurring faults require a suitable reference for this comparison.

\bibliographystyle{tmlr}
\bibliography{references}
\appendix
\renewcommand{\theHfigure}{A\arabic{figure}}
\renewcommand{\theHtable}{A\arabic{table}}

\renewcommand{\thefigure}{A\arabic{figure}}
\setcounter{figure}{0}
\renewcommand{\thetable}{A\arabic{table}}
\setcounter{table}{0}
\section{Supplementary Results and Experimental Details}\label{appendix-a-supplementary-results}

The appendix supplies the construction and protocols needed to reproduce the framework comparisons: action sampling and prompts, component eligibility and search, database component results, simulator details, and the held-out execution-preview protocol.

\subsection{Component eligibility, search, and scoring}\label{a-component-search}

\textbf{Eligibility.} The component analysis starts from all 140 demand episodes in the difficulty study. Of the 117 actions that choose demand scaling, three have empty target sets. The remaining 114 enter the two-factor comparison. The search study includes 11 fixed-evidence and 43 feedback actions from eight fault cases, comprising 28 distinct state--target combinations. In both studies, eligibility requires the demand-scaling operation and a nonempty target set and is determined before evaluating replacement outcomes.

\textbf{Scoring.} The recovery rate is computed from loss on the fixed set of observed links in Equation~\ref{eq:traffic-loss}. In the overall numerical-repair evaluations, scoring includes no-ops and incorrect diagnoses. The recovery rate is unbounded below. Sensor exclusion changes the diagnostic panel and leaves the numerical recovery rate unchanged when it leaves simulated flow unchanged.

Exact target precision and recall compare selected and affected OD pairs. Positive-demand recall restricts this comparison to pairs with nonzero demand, for which scaling changes the state.

\textbf{Symmetric component allocation.} For summaries across actions, we average each replacement effect over the two possible settings of its partner component. The resulting Shapley allocation \citep{shapley1953value} is
\begin{equation}
\begin{aligned}
\Delta_S &= \tfrac12\{(R_{10}-R_{00})+(R_{11}-R_{01})\},\\
\Delta_u &= \tfrac12\{(R_{01}-R_{00})+(R_{11}-R_{10})\}.
\end{aligned}
\end{equation}
The effects sum to $R_{11}-R_{00}$. Each combines the direct replacement effect with half the interaction in Equation~\ref{eq:component-interaction}. This allocation and the effect of replacement with proposed targets answer different questions. For one DeepSeek-Flash database action in Figure~\ref{fig:database-outcomes}, $(R_{00},R_{10},R_{01},R_{11})=(0.055,0.051,-0.008,1.000)$. Update replacement with proposed targets changes recovery by $-0.063$, while its symmetric allocation is $\Delta_u=0.443$ because the reference update helps after target correction.

\textbf{Component effects across difficulty conditions.} Table~\ref{tab:component-strata} groups the 114 actions by target accuracy to show how the remaining repair opportunity changes with selection quality. Table~\ref{tab:ladder-components-model} reports the component effects for each model.

\begin{table}[ht]
\caption{\textbf{Component effects by target accuracy across 114 demand actions.}}
\label{tab:component-strata}
\centering
\begin{tabular}{lrrr}
\toprule
Target F1 & Actions & $\Delta_S$ & $\Delta_u$ \\
\midrule
Below 0.5 & 60 & 0.583 & 0.247 \\
0.5 to below 0.9 & 18 & 0.449 & 0.224 \\
0.9 and above & 36 & 0.000 & 0.063 \\
\bottomrule
\end{tabular}
\end{table}

\begin{table}[ht]
\caption{\textbf{Component effects by model in the difficulty study.} Effects are mean marginal changes in recovery rate under the two-factor comparison.}
\label{tab:ladder-components-model}
\centering
\begin{tabular}{lrrrr}
\toprule
Model & Eligible actions & Agent $R$ & $\Delta_S$ & $\Delta_u$ \\
\midrule
K3 & 23 & 0.664 & 0.274 & 0.062 \\
DeepSeek-Flash & 23 & 0.345 & 0.487 & 0.168 \\
GLM-5.3 & 23 & 0.575 & 0.229 & 0.196 \\
GLM-Flash & 23 & 0.465 & 0.378 & 0.157 \\
MiMo-Pro & 22 & 0.122 & 0.527 & 0.351 \\
\bottomrule
\end{tabular}
\end{table}

\textbf{Fixed-target search.} Each search evaluates 25 logarithmically spaced multipliers over $[0.200,8.000]$ and three interior refinement points around the best grid point. The mean recovery rate is 0.437 when actions are weighted equally and 0.421 when the 28 distinct state--target combinations are weighted equally. Table~\ref{tab:complete-component-replacement} reports the resulting means.

\begin{table}[ht]
\caption{\textbf{Component replacements and fixed-target search.} Each row specifies which components of the proposed action are replaced. The search row holds the proposed targets fixed and searches the legal multiplier range.}
\label{tab:complete-component-replacement}
\centering
\begin{tabular}{lrrrr}
\toprule
Execution & Mean $R$ & Median $R$ & Max $R$ & $R\geq 0.9$ \\
\midrule
Proposed action & 0.113 & 0.117 & 0.291 & 0/54 \\
Replace update & 0.206 & 0.176 & 0.432 & 0/54 \\
Fixed-target search & 0.437 & 0.450 & 0.696 & 0/54 \\
Replace targets & 0.794 & 0.867 & 1.000 & 22/54 \\
Reference action & 1.000 & 1.000 & 1.000 & 54/54 \\
\bottomrule
\end{tabular}
\end{table}

Target replacement has the larger effect across both evidence sources. In the 11 fixed-evidence actions, the mean recovery rate is 0.070 for the proposed action, 0.790 with reference targets, and 0.110 with the reference update. The target-set effect exceeds the multiplier effect in 11/11 actions. In the 43 feedback-derived actions, the corresponding recovery rates are 0.124, 0.794, and 0.230, with the same ordering in 43/43 actions. The ordering also holds separately for the initial, second-chance, scalar-feedback, and localized-feedback branches.

The absolute paired difference between target replacement and fixed-target search exceeds 0.01 in every action.

\textbf{Normalization sensitivity.} Numerical route assignment can leave a small discrepancy after the traffic inputs are restored. The recovery-rate denominator retains this residual reference loss. The pooled recovery rate divides summed loss reductions by summed reference improvements. Across the sparse-target actions, it is 0.362 for fixed-target search and 0.801 for target replacement. Median recovery rates are 0.450 and 0.867.

Pooled component effects preserve the target-dominant ordering in the low-F1 group and the update-dominant ordering in the high-F1 group. Weighting matters in the intermediate group. Equal-action target and update effects are 0.449 and 0.224, while pooled effects are 0.086 and 0.459. The mean recovery rate gives each action equal weight. The pooled recovery rate gives greater weight to larger reference loss reductions.

\subsection{Fault recognition, target accuracy, and outcome profiles}\label{difficulty-profile-results}
The profiles below show how fault recognition, target accuracy, and recovery rate vary with fault magnitude and spatial distribution. The four-action comparisons in the main text attribute repair outcomes to target and update decisions.

\begin{figure}[!htb]
\centering
\includegraphics[width=\textwidth]{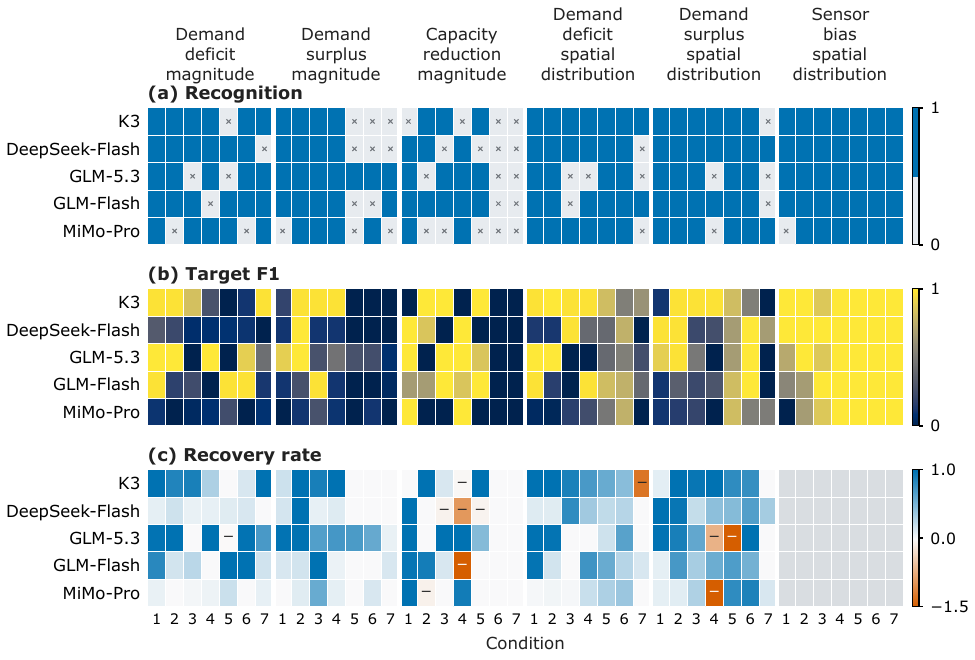}
\caption{\textbf{Fault recognition, target accuracy, and recovery rate form different difficulty profiles.} Each block contains seven conditions varying fault magnitude, the number and arrangement of affected OD pairs, or the share of affected sensors. Panels show (a) fault recognition, with blue cells denoting correct classification and crosses marking incorrect ones, (b) target F1, and (c) recovery rate with colors saturated outside $[-1.5,1]$ and minus signs for harmful actions. The gray block in (c) denotes sensor repair, evaluated by target accuracy. Condition settings are listed in Appendix~\ref{a7-tool-catalogue-and-benchmark-difficulty-details}.}
\label{fig:difficulty-ladder-results}
\end{figure}

\subsection{Instance specifications}\label{a4-instance-specifications}
\subsubsection{Database instances and evaluation}\label{a-database}

The database is generated at TPC-H scale factor 0.100. Before fault injection, product prices receive record-key-dependent multipliers between 1.000 and 1.200, and transaction amounts are recomputed from prices and quantities. An adjustment rate $d$ becomes $0.950d+0.008+\delta$, where $\delta\in[-0.001,0.001]$ is determined by record keys. Rates are clipped to $[0.001,0.099]$. Identifiers use opaque aliases, group sizes appear as rounded shares, and editable values retain ten decimal places.

Database requests use K3 (\texttt{moonshotai/kimi-k3}) and GLM-5.3 (\texttt{z-ai/glm-5.3}) through OpenRouter, and DeepSeek-Flash (\texttt{deepseek-flash}) through the native DeepSeek endpoint. Output limits are 49,152 tokens with a 1,800-second request deadline. GLM-5.3 uses low reasoning, and DeepSeek-Flash disables thinking. Both evidence conditions share each model's settings. The three generations for each model--fault--evidence combination use identical model-visible messages. Every valid action enters the four component executions.

\begin{figure}[ht]
\centering
\includegraphics[width=\textwidth]{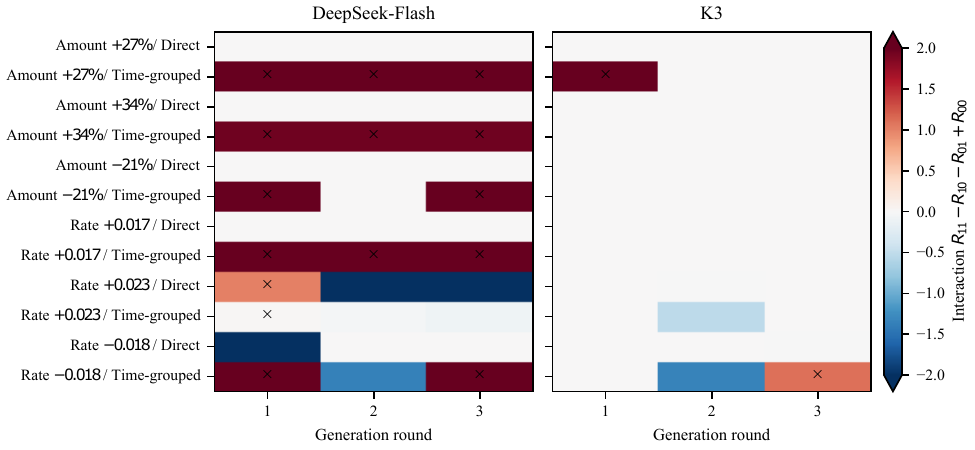}
\caption{\textbf{Component interactions and update-replacement harm for DeepSeek-Flash and K3.} Each column is one generation round and each row one fault under one evidence condition. Color shows the interaction $R_{11}-R_{10}-R_{01}+R_{00}$, saturated at $\pm2$. A cross marks an action whose recovery falls by more than 0.01 when only the update is replaced. GLM-5.3 has no harmed action, so its panel is omitted.}
\label{fig:database-outcomes}
\end{figure}

Table~\ref{tab:database-repeat-summary} separates replacement outcomes by model and round. Figure~\ref{fig:database-outcomes} shows the action-level outcomes of the two models with harmed actions.

The six faults affect intersections of shipping mode, handling instruction, and return status (Table~\ref{tab:database-faults}). A target predicate joins categorical conditions by conjunction within each of up to two disjunctive clauses. An update selects a field and applies a shared multiplication or addition.

\begin{table}[ht]
\centering
\caption{\textbf{Database fault definitions.} Each intersection selects one category of each named attribute. Amount changes are multiplicative, and adjustment-rate changes are additive.}
\label{tab:database-faults}
\begin{tabular}{lll}
\toprule
Fault & Affected intersection & Injected change \\
\midrule
Amount $+27\%$ & Shipping mode and return status & Multiply amount by 1.270 \\
Rate $+0.023$ & Shipping mode and return status & Add 0.023 to adjustment rate \\
Amount $-21\%$ & Handling instruction and return status & Multiply amount by 0.790 \\
Rate $-0.018$ & Handling instruction and return status & Subtract 0.018 from adjustment rate \\
Amount $+34\%$ & Shipping mode and handling instruction & Multiply amount by 1.340 \\
Rate $+0.017$ & Shipping mode and handling instruction & Add 0.017 to adjustment rate \\
\bottomrule
\end{tabular}
\end{table}

Each observation panel reports total amount, $\sum_i a_i$, the sum of adjustment rates, $\sum_i d_i$, and total adjusted amount, $\sum_i a_i(1-d_i)$, where $a_i$ and $d_i$ denote amount and adjustment rate. It contains the global residual and up to 35 groups with the largest residuals. Direct evidence includes aggregates for the affected combination of categories. Time-grouped evidence replaces that view with a temporal grouping, which also changes the groups entering the truncated panel. Both conditions share instructions, generation settings, action grammar, executor, and scoring queries.

\textbf{Scoring workload.} The scoring workload computes these three aggregates for every group. Groups comprise the table, each category of shipping mode, handling instruction, return status, line status, and shipping year, and each observed combination of shipping mode with year, handling instruction with return status, shipping mode with return status, shipping mode with line status, and shipping mode with handling instruction and year. These queries produce 316 groups and 948 aggregate values indexed by $Q$. Each aggregate contributes one equally weighted loss term. Both evidence conditions and all component executions use this workload.

For aggregate $k\in Q$, let $z_k$ be its current value and $y_k$ its healthy value. Loss averages $(z_k-y_k)^2/(|y_k|+\epsilon_{\mathrm{database}})^2$ over $Q$, with $\epsilon_{\mathrm{database}}=10^{-12}$. Every inverse repair restores the stored values and achieves zero loss. The recovery rate therefore equals the fraction of pre-repair loss removed.

\FloatBarrier
\textbf{Actions in Table~\ref{tab:database-components}.} The DeepSeek-Flash example concerns a 34\% amount increase within the intersection of one shipping-mode category and one handling-instruction category. The model selects the entire shipping-mode category and multiplies amounts by 0.923. The reference multiplier is 0.746. Its 85,988 selected rows include all 21,233 affected rows and 64,755 unaffected rows. This is the time-grouped-evidence proposal in the first repetition of the amount-increase condition in Figure~\ref{fig:database-outcomes}. The GLM-5.3 example concerns a 21\% amount decrease within a handling--return-status intersection, using time-grouped evidence in the first repetition. Its target set is exact. Its multiplier is 1.724, and the reference multiplier is 1.266. Displayed multipliers are rounded to three decimal places.


\begin{table}[H]
\centering
\caption{\textbf{Database replacement outcomes across three generation rounds.} Each model contributes 12 actions per round (six faults, two evidence conditions). Excess means that selected rows include unaffected rows. Harm means $R_{01}-R_{00}<-0.01$. The final columns describe the recovery-rate effect of update replacement over all 12 actions.}
\label{tab:database-repeat-summary}
\small
\begin{tabular}{llrrrrr}
\toprule
Model & Round & Excess & Harm/all & Harm/excess & Mean $\Delta R$ & Median $\Delta R$ \\
\midrule
DeepSeek-Flash & 1 & 8/12 & 7/12 & 7/8 & -0.570 & -0.042 \\
 & 2 & 6/12 & 3/12 & 3/6 & 0.137 & 0.001 \\
 & 3 & 7/12 & 5/12 & 5/7 & 7.192 & 0.000 \\
K3 & 1 & 1/12 & 1/12 & 1/1 & -0.112 & 0.000 \\
 & 2 & 2/12 & 0/12 & 0/2 & 0.449 & 0.000 \\
 & 3 & 1/12 & 1/12 & 1/1 & 0.034 & 0.000 \\
GLM-5.3 & 1 & 1/12 & 0/12 & 0/1 & 0.682 & 0.170 \\
 & 2 & 0/12 & 0/12 & --- & 1.037 & 0.000 \\
 & 3 & 0/12 & 0/12 & --- & 9.013 & 0.000 \\
\bottomrule
\end{tabular}
\end{table}
Large positive mean effects can coexist with frequent harm because correcting an update can rescue a few actions with very large negative original recovery rates. The mean and harm count describe different aspects of the distribution.
\FloatBarrier

\FloatBarrier

\FloatBarrier

\FloatBarrier

\subsubsection{Traffic instances}

The targeted traffic probes and interface comparisons use the following instances:

\begin{itemize}
\item
  A 22\% capacity reduction uses injected multiplier 0.781.
\item
  A 27\% demand surplus uses injected multiplier 1.268.
\item
  The two-origin surplus case is a 33\% demand surplus with multiplier 1.330 and 48 planted rows across two origins.
\item
  The five-origin surplus case is a 21\% demand surplus with multiplier 1.210 and 120 planted rows across five origins.
\item
  The two-origin deficit case is a 38\% demand deficit with multiplier 0.620 and 48 planted rows across two origins.
\item
  The three-origin deficit case is a 26\% demand deficit with multiplier 0.740 and 72 planted rows across three origins.
\end{itemize}

Each reference action restores the generating demand or capacity parameters. The remaining post-repair discrepancy is at the healthy numerical floor. Each fault loss exceeds 100 times that floor. Affected OD-pair sets come from seeded candidates near the median total demand of those sets, subject to an origin-set Jaccard similarity threshold of 0.30 against other evaluated sets.

\subsection{Tool catalogue and testbed details}\label{a7-tool-catalogue-and-benchmark-difficulty-details}

\textbf{Network implementation.} The simulator uses the Sioux Falls traffic-assignment benchmark \citep{leblanc1975traffic}. It has 24 zones, 76 directed links, and a $24\times24$ OD matrix. Healthy reference flows define a fixed 76-link observation panel. Anonymous variants replace identifiers and perturb healthy demand before regenerating those observations.

\textbf{Traffic notation and travel cost.} An origin--destination (OD) pair specifies where trips start and end. A directed link is one road connection along a route. The OD matrix lists the demand for each pair. Assigned link flow $v_a$ is the number of trips routed through link $a$. The volume-to-capacity ratio $v/c$ divides assigned flow by link capacity. The Bureau of Public Roads (BPR) function gives the travel time on link $a$ \citep{fhwa2022multiresolution}:
\begin{equation}
 t_a(v_a)=t_a^0\left[1+0.150\left(\frac{v_a}{c_a}\right)^4\right].
\end{equation}
Here $t_a^0$ is free-flow travel time and $c_a$ is capacity. Both start from the Sioux Falls network values. Instance transformations and repair actions determine the current capacities and demand.

\textbf{Route assignment.} User equilibrium means that every used route has minimum travel time for its OD pair under the resulting link flows. The Frank--Wolfe algorithm computes this assignment \citep{leblanc1975traffic}. Each iteration assigns all demand for each OD pair to a shortest route at the current travel times. This all-or-nothing assignment supplies a search direction. A line search then blends it with the current flows by minimizing $\sum_a\int_0^{v_a}t_a(w)\,dw$. Shortest paths and tie handling are deterministic. The line-search tolerance is $10^{-12}$.

Let $d_{od}$ denote demand and $\ell_{od}(v)$ the shortest-route travel time at flows $v$. The relative gap is
\begin{equation}
 g(v)=\frac{\sum_a v_a t_a(v_a)-\sum_{od}d_{od}\ell_{od}(v)}{\sum_a v_a t_a(v_a)}.
\end{equation}
The model's observations are fixed before converged outcome scoring. Assignments after demand or capacity edits target $g\leq10^{-5}$. Outcomes in the 11-model evaluation and difficulty study allow up to 100,000 iterations. Fixed-target search allows up to 200,000 where needed. Each action is applied to a fresh copy of its pre-action state before route assignment.

\textbf{Observation loss.} For simulated flow $q$ and reference flow $y$, the implementation computes the GEH statistic \citep{transportscotland2008geh} as
\begin{equation}
  \operatorname{GEH}(q,y)=\sqrt{\frac{2(q-y)^2}{q+y+\epsilon_{\mathrm{traffic}}}},
  \end{equation}
where $\epsilon_{\mathrm{traffic}}=10^{-9}$ prevents division by zero. The discrepancy is zero when both counts are zero. Loss $J$ is the mean squared GEH over the fixed observation panel. The recovery rate in Equation~\ref{eq:recovery} normalizes improvement in $J$ by the reference action's loss reduction.

\subsubsection{Tool catalogue}\label{tool-catalogue}

The interactive protocol offers read-only tools that expose complementary projections of the system:

\begingroup
\raggedright
\begin{itemize}
\item
  \path{demand_profile}: total demand, OD count, trip-length distribution, top origins, destinations, and OD pairs.
\item
  \path{sensor_profile}: count distributions and road-class summaries.
\item
  \path{road_class_mismatch}: count-share and network-length discrepancies by road class.
\item
  \path{zone_outliers}: production or attraction tails by zone.
\item
  \path{demand_sensor_consistency}: production-to-incident-count consistency and extreme zones.
\item
  \path{congestion_scan}: $v/c$ and BPR time-ratio summaries.
\item
  \path{residual_geh_by_class}: per-sensor and per-class simulated-versus-observed residuals.
\item
  \path{link_residual_scan}: capacity, assigned flow, $v/c$, observed count, simulated-to-observed ratio, and GEH.
\item
  \path{support_coverage}: the volume and count of OD pairs sensitive to integer rounding.
\end{itemize}
\endgroup

The interactive protocol returns complete tool results for the requested scope. Targeted probes either supply the correct class during interaction or present a fixed final evidence state. Capacity probes additionally supply target identifiers, leaving the agent to choose the multiplier.

\subsubsection{Difficulty grids and fault-class visibility}\label{difficulty-grids-and-certification}

The difficulty-study instances use seed 42. Fault magnitude is parameterized by $a=|\ln m|$, with larger values producing stronger faults. The spatial-distribution series vary the number and arrangement of affected OD pairs or the set of affected sensors. The evaluated cases include both demand directions, capacity reductions, sensor bias, and a healthy control.

A fault-class visibility test uses the same summaries available to the agent. These include directional residuals, high-GEH tails and their $v/c$ values, demand--sensor consistency, spatial concentration, and similarity in simulated-to-observed flow ratios among high-residual sensors. A healthy system establishes the numerical solver noise. The test selects settings where the injected class is distinguishable within the benchmark's finite hypothesis set. Exact target metrics quantify localization within the class.

The seven conditions in each series use the following settings. For demand-deficit magnitude, $a$ takes values $1.28, 0.96, 0.64, 0.48, 0.32, 0.24,$ and $0.16$. Demand-surplus and capacity-reduction magnitude use $1.28, 0.64, 0.32, 0.24, 0.16, 0.12,$ and $0.08$. The demand spatial-distribution series use affected origins in contiguous groups of 7, 5, 3, and 2 zones, followed by 16-pair and 8-pair sets that share a single origin and a dispersed set of 10 OD pairs. The sensor-bias spatial-distribution series scales the affected share of sensors through $1, 0.75, 0.50, 0.25, 0.125, 0.08,$ and $0.04$. The magnitude is fixed at $a=0.64$ within each spatial-distribution series.

\subsection{Fixed-evidence evaluation}\label{a1-fixed-evidence-evaluation}

The fixed diagnostic state contains summaries of trips and sensor readings, discrepancies by road type, unusually large or small origin and destination totals, and comparisons between trip demand and road counts. It also includes traffic-load and solver checks, residuals for all 76 directed links, and information about trip counts sensitive to integer rounding. Outputs are produced once per case, anonymized, compacted, and reused across models. The model infers the fault class and targets. The action protocol uses JavaScript Object Notation (JSON) objects for both tool calls and final actions. Appendix~\ref{a2-prompts-and-endpoints} records the literal shared instruction.

\begin{table}[ht]
\caption{\textbf{Conditions in the fixed-evidence evaluation.} Each state is checked for fault-class visibility and recovery rate under the reference action before model execution.}
\label{tab:fixed-evidence-cases}
\centering
\begin{tabular}{llll}
\toprule
Case & Condition & Injected multiplier & Planted targets \\
\midrule
Broad deficit & Demand deficit & 0.616 & 168 OD pairs \\
Broad surplus & Demand surplus & 1.268 & 168 OD pairs \\
Capacity reduction & Capacity reduction & 0.781 & 3 links \\
Localized deficit & Demand deficit & 0.532 & 48 OD pairs \\
Localized surplus & Demand surplus & 1.878 & 48 OD pairs \\
Sensor bias & Sensor bias & 1.878 & 6 sensors \\
Healthy control & Healthy control & --- & no targets \\
\bottomrule
\end{tabular}
\end{table}

K3, GLM-Flash, DeepSeek-Flash, GLM-5.3, and MiMo-Pro each supply one final valid action for each of the seven conditions, yielding 35 actions. Table~\ref{tab:fixed-evidence-cases} gives the injected magnitudes and planted targets of the seven conditions. Requests are retried after request or output-format failures. All 35 actions enter fault recognition and validity results, and all 25 numerical actions enter recovery-rate results. Table~\ref{tab:fixed-evidence} reports the model-level summary.

\begin{table}[!htbp]
\caption{\textbf{Fixed-evidence results by model.} Overall fault recognition covers all seven conditions. Numerical recognition and recovery rate cover the five numerical conditions. The mean recovery rate includes no-ops and incorrect diagnoses.}
\label{tab:fixed-evidence}
\centering
\begin{tabular}{lrrrrr}
\toprule
Model & Recognized & \shortstack{Recognized\\numerical cases} & Mean $R$ & $R\geq 0.9$ & $R<0$ \\
\midrule
K3 & 4/7 & 2/5 & 0.023 & 0/5 & 0/5 \\
GLM-Flash & 4/7 & 2/5 & 0.036 & 0/5 & 0/5 \\
DeepSeek-Flash & 2/7 & 0/5 & 0.000 & 0/5 & 0/5 \\
GLM-5.3 & 6/7 & 4/5 & 0.083 & 0/5 & 0/5 \\
MiMo-Pro & 5/7 & 3/5 & 0.015 & 0/5 & 0/5 \\
\bottomrule
\end{tabular}
\end{table}

\textbf{Observations.} In the fixed-evidence broad-deficit condition, MiMo-Pro, K3, GLM-Flash, and GLM-5.3 all choose multiplier 1.500. They select one, five, 14, and 20 OD pairs, respectively, with recovery rates of 0.004, 0.023, 0.141, and 0.224. All five evaluated models select exactly the six biased sensors and abstain on the healthy control.

\textbf{Target visibility.} Across four fixed demand conditions, every affected origin and both identifiers of every affected OD pair appear in the evidence. Explicitly displayed demand-table rows cover 1.8\% of affected OD pairs on average. Across 140 interactive demand episodes, the evidence contains 99.9\% of affected origin identifiers and both identifiers for 98.8\% of affected pairs. Explicit demand-table row coverage is 11.1\%. Agents can express the targets using visible identifiers, and complete selection requires inferring membership beyond the displayed rows. The interface experiment supplies the full positive-demand table.

\FloatBarrier

\begin{table}[!htbp]
\caption{\textbf{Replacing the update can improve or worsen a repair.} Recovery rates for two database repairs. DeepSeek-Flash selects a broad group (target F1 0.396), and GLM-5.3 selects the exact affected set. Each column compares four executions of one proposal.}
\label{tab:database-components}
\begin{center}
\begin{tabular}{lrr}
\toprule
Action & DeepSeek-Flash & GLM-5.3 \\
\midrule
Proposed action & 0.427 & $-1.973$ \\
Replace targets & 0.517 & $-1.973$ \\
Replace update & $-1.032$ & 1.000 \\
Reference action & 1.000 & 1.000 \\
\bottomrule
\end{tabular}

\end{center}
\end{table}

\subsection{Feedback protocol}\label{a3-feedback-sampling-validity-and-sequencing}

The feedback evaluation tests K3, GLM-Flash, and DeepSeek-Flash on four demand cases with fixed generation rules, prompts, branch conditions, and model settings. Each revision request includes the original evidence and the initial action. Scalar feedback gives the traffic loss before and after that action, indicates whether it decreased, and states that lower loss is better. Localized feedback adds post-action demand--sensor consistency summaries and a residual scan of the 20 highest-residual links. The second-chance condition asks the model to reconsider its initial action using the original evidence. All 12 model--fault cells contain a valid initial action and scalar-feedback revision. Ten also contain a localized-feedback revision. Nine contain a second-chance request for the same case using the original evidence.

All 12 scalar-feedback revisions retain the exact initial targets. Five improve the recovery rate, and seven leave it unchanged. Seven of ten localized-feedback revisions retain the initial targets, and all ten preserve or improve the initial coverage of affected records.

\subsection{Prompts, endpoint identifiers, and decoding}\label{a2-prompts-and-endpoints}

Fixed-evidence requests use the shared final-action instruction below after receiving the standardized diagnostic state. A zone is a location at which trips start or end. A road class groups links of the same type. Congestion summaries describe traffic load relative to road capacity, including the volume-to-capacity ratio ($v/c$). Equilibrium refers to the route assignment reached when travelers have reached minimum travel time on their used routes. Request traces preserve the exact instructions and model-visible states.

\begin{promptbox}{Prompt A1: Fixed-evidence traffic repair}
\begin{Verbatim}[breaklines=true,breakanywhere=true,breaksymbolleft={},breaksymbolright={}]
You are a traffic-calibration diagnostician performing differential diagnosis on a planted-or-absent data fault. Reply with ONE JSON object matching the action schema -- no prose, no markdown.
Actions:
  {"action": "attribute", "attribution": {<Attribution fields>}}
Attribution fields: fault_class, intervention_type, intervention_magnitude, confidence, rationale, target_zones, target_sensor_ids, target_od_pairs, road_class.
Fault classes (use exactly):
  - demand_deficit: A localized footprint of OD entries is too low. Look for simulation under-predicting observed counts around the affected origins; name the OD pairs and use scale_od with a factor above 1.
  - demand_surplus: A localized footprint of OD entries is too high. Look for simulation over-predicting observed counts around the affected origins; name the OD pairs and use scale_od with a factor below 1.
  - sensor_bias: One or more sensors' observed counts are systematically scaled. The affected scope may be a whole road class, a subset of one class, or an isolated sensor. Only sim-vs-observed residuals identify it -- data profiles cannot. Name the sensors to exclude with downweight_sensors; its magnitude field is ignored.
  - capacity_cut: Supply-side lane or capacity drop on a few high-flow links. Demand profiles stay healthy; identify via residuals plus v/c (congestion_scan).
  - none: No planted fault. Healthy baselines can contain high-v/c links or small residuals; use none when no listed fault is supported.
Intervention types (use exactly): scale_od, downweight_sensors, restore_capacity, escalate, none
intervention_magnitude is the factor multiplied onto the named rows of the CURRENT demand or link table. Sensor exclusion ignores this field. The harness does not look up a hidden true table. Infer the factor from the observed residuals. For scaling actions, supply a finite number in [0.2, 8.0]. For none, escalate, and downweight_sensors, use 1.0; missing/null magnitudes on these non-scale actions are normalized to 1.0. confidence is a finite number in [0, 1]; low/medium/high are accepted as representation aliases for 0.25/0.5/0.75, not calibrated probabilities. For scale_od, set target_od_pairs to the affected [origin, dest] pairs (two-element lists of the identifiers returned by tools). target_zones is descriptive only; it does not select OD rows. A healthy equilibrium may be congested; use none rather than treating congestion as a planted fault.
Protocol:
- All residuals use the assignment under the current demand and network.
- Mild baseline anomalies exist even in healthy data. `none` is a legitimate answer -- do not force a fault when residuals are ordinary.
- When you emit action=attribute, rationale MUST cite specific metric values from the observations (not generic impressions), including why you chose that intervention_magnitude.
- Each observation result is represented as {summary, flags, result_preview}. summary, flags, and result_preview are complete; result_preview omits filesystem provenance.
For this request, return exactly one textual JSON object containing exactly one top-level action. Do not append another action, an empty object, prose, or markdown. The request ends after that object.
Before emitting the final action, independently check that every action field is supported by the available evidence and that the action is internally consistent. Do not assume unavailable values.
Identifiers are anonymous aliases. Use only identifiers supplied by tools. Healthy numerical parameters vary legitimately between instances. All residuals use the current demand/network assignment.
Private attribution field types: target_zones and target_sensor_ids must be JSON arrays of integer identifiers, never prose strings. Use [] when a target field is unused. target_od_pairs must be an array of two-integer arrays [origin, destination]. The phrase 'target_zones is descriptive only' means this integer array is metadata and does not select demand rows; put all prose descriptions in rationale. road_class is a string or null. For none, all three target arrays must be empty. For restore_capacity or downweight_sensors, target_sensor_ids names the affected sensors. For scale_od, target_od_pairs names the affected demand rows.
\end{Verbatim}
\end{promptbox}

\textbf{Database instruction.} The box below presents the fixed instruction using field meanings. The user message supplies the editable fields, allowed predicate values, aggregate residuals, and their definitions. Each request returns one repair action. Appendix~\ref{a-database} specifies the evidence panels and scoring workload.

\begin{promptbox}{Prompt A2: Database repair instructions (semantic rendering)}
\begin{Verbatim}[breaklines=true,breakanywhere=true,breaksymbolleft={},breaksymbolright={}]
You repair a numerical inconsistency in an anonymous relational reporting table. A healthy reference reporting workload and the current database disagree. The user supplies aggregate relative errors but never supplies the planted row predicate or the correct inverse update. Table, field, category, and case names are opaque aliases; infer the repair only from the supplied evidence.

Return one structured action containing the selected records, a numerical update, and a rationale.

Select records through one or two clauses. A row qualifies if either clause holds; within a clause, all specified conditions must hold. Conditions concern shipping mode, handling instruction, return status, line status, and shipping year. Each condition lists allowed categories; an empty list leaves that attribute unrestricted. At least one attribute must be restricted in each clause. Use only supplied categories.

Specify the field (amount or adjustment rate), the operator (multiplication or addition), and the factor or signed increment applied to the current value. Infer an update that reverses the discrepancy. Multipliers must be in [0.2,5.0], additive updates in [-0.2,0.2].

The executor applies the update once to every row selected by the predicate. Aggregate evidence can identify a conjunction even when its marginal groups also show residuals. Choose the complete affected support while avoiding unaffected rows. Return no markdown and no additional fields.
\end{Verbatim}
\end{promptbox}

For the fixed-evidence study, Table~\ref{tab:model-request-record} reports application programming interface (API) identifiers, evaluation dates, and decoding settings. Table~\ref{tab:survey-endpoints} lists the endpoints of the 11-model evaluation, and Table~\ref{tab:provider-routing} the provider that served each fixed-evidence condition. DeepSeek-Flash requests used the \texttt{deepseek-flash} alias, JSON-object output, and provider-default temperature and top-$p$.

The interactive adapter executes the first parsed JSON action in each response, with one tool call per turn. Final repair actions undergo field, identifier, and range validation. Fixed-evidence requests use a single-object parser. The 11-model evaluation allows 25 tool calls and 49,152 output tokens per request, with high reasoning effort and episode limits of 900--3,600 seconds.

\begin{table}[ht]
\caption{\textbf{Endpoint identifiers for the 11-model traffic evaluation.} }
\label{tab:survey-endpoints}
\centering
\begin{tabular}{lll}
\toprule
Short name & Endpoint identifier & Evaluation date (2026) \\
\midrule
Astra & \texttt{openai/gpt-6-astra} & Sep 12 \\
Sol & \texttt{openai/gpt-5.6-sol} & Sep 12 \\
Terra & \texttt{openai/gpt-5.6-terra} & Sep 12 \\
Opus & \texttt{anthropic/claude-opus-4.8} & Sep 12 \\
Sonnet & \texttt{anthropic/claude-sonnet-4.6} & Sep 12 \\
Gemini & \texttt{google/gemini-3.8-flash} & Sep 12 \\
K3 & \texttt{kimi-k3} & Sep 12 \\
DeepSeek-Flash & \texttt{deepseek-flash} & Oct 17 \\
GLM-5.3 & \texttt{z-ai/glm-5.3} & Sep 12 \\
GLM-Flash & \texttt{z-ai/glm-5.3-flash} & Sep 12 \\
MiMo-Pro & \texttt{xiaomi/mimo-v2.5-pro} & Sep 12 \\
\bottomrule
\end{tabular}
\end{table}

\begin{table}[ht]
\caption{\textbf{Endpoint and decoding record for the fixed-evidence evaluation.} $^\dagger$GLM-5.3's broad surplus condition uses 8,192 tokens and low reasoning. All other GLM-5.3 cells use 16,384 tokens and high reasoning.}
\label{tab:model-request-record}
\centering
\begin{tabular}{lllll}
\toprule
Model & Endpoint identifier & Date (2026) & Max tokens & Reasoning \\
\midrule
K3 & \texttt{kimi-k3} & Sep 14 & 16,384 & unspecified \\
DeepSeek-Flash & \texttt{deepseek-flash} & Oct 17 & 8,192 & disabled \\
GLM-Flash & \texttt{z-ai/glm-5.3-flash} & Sep 14 & 16,384 & high \\
GLM-5.3 & \texttt{z-ai/glm-5.3} & Sep 15 & 16,384$^\dagger$ & high$^\dagger$ \\
MiMo-Pro & \texttt{xiaomi/mimo-v2.5-pro} & Sep 15 & 16,384 & unspecified \\
\bottomrule
\end{tabular}
\end{table}

\begin{table}[ht]
\caption{\textbf{Provider routes for the fixed-evidence evaluation.}}
\label{tab:provider-routing}
\centering
\begin{tabular}{@{}lll@{}}
\toprule
Model & Conditions & Returned provider \\
\midrule
K3 & All & native endpoint \\
DeepSeek-Flash & All & native endpoint \\
GLM-Flash & All & StreamLake \\
GLM-5.3 & Broad deficit, broad surplus & DeepInfra \\
GLM-5.3 & Capacity reduction, localized deficit, sensor bias & Reka \\
GLM-5.3 & Localized surplus, healthy & Morph \\
MiMo-Pro & Broad deficit, broad surplus & Xiaomi \\
MiMo-Pro & Capacity reduction, localized surplus, sensor bias & GMICloud \\
MiMo-Pro & Localized deficit, healthy & AtlasCloud \\
\bottomrule
\end{tabular}
\end{table}

All five models in the difficulty study used JSON-object output, 25 tool calls, a 3,600-second request deadline, and a 14,400-second episode ceiling. DeepSeek-Flash used 65,536 output tokens, and the other models used 49,152. High reasoning was requested where supported. Table~\ref{tab:ladder-endpoints} records the model routes used uniformly across all 43 cells for each model.

\begin{table}[ht]
\caption{\textbf{Endpoint and decoding record for the difficulty study.} Each row uses one gateway and configuration over the complete 43-condition matrix.}
\label{tab:ladder-endpoints}
\centering
\begin{tabular}{lllll}
\toprule
Model & Endpoint identifier & Gateway & Max tokens & Reasoning \\
\midrule
K3 & \texttt{moonshotai/kimi-k3} & OpenRouter & 49,152 & high \\
DeepSeek-Flash & \texttt{deepseek-flash} & native endpoint & 65,536 & high \\
GLM-5.3 & \texttt{z-ai/glm-5.3} & OpenRouter & 49,152 & high \\
GLM-Flash & \texttt{z-ai/glm-5.3-flash} & OpenRouter & 49,152 & high \\
MiMo-Pro & \texttt{xiaomi/mimo-v2.5-pro} & OpenRouter & 49,152 & provider default \\
\bottomrule
\end{tabular}
\end{table}

\FloatBarrier
\begin{table}[!b]
\centering
\caption{\textbf{Four combinations of two GLM-5.3 proposals.} Three-origin demand deficit, repetition 2, selection panel A. Target counts are positive-demand OD pairs. The lowest selection-panel loss determines the preview choice.}
\label{tab:preview-crossing-example}
\begin{tabular}{lrrrr}
\toprule
Candidate & Targets & Multiplier & Selection loss & Held-out $R$ \\
\midrule
Record targets, record update & 5 & 1.250 & 40.224 & 0.046 \\
Grouped targets, grouped update & 103 & 1.450 & 71.762 & 0.218 \\
Record targets, grouped update & 5 & 1.450 & 35.856 & 0.079 \\
Grouped targets, record update & 103 & 1.250 & 13.749 & 0.680 \\
\bottomrule
\end{tabular}
\end{table}

\subsection{Execution-preview protocol}\label{a3b-target-representation}

For each model, fault, and panel direction, both interfaces receive the same selection-panel diagnostic state and 528-row positive-demand table. Diagnostic summaries use the 38 selection-panel sensors. In the record-level interface, \texttt{target\_od\_pairs} selects OD pairs, and \texttt{target\_zones} is descriptive metadata. In the grouped interface, the executor expands \texttt{target\_origins} to the corresponding OD pairs with positive demand in the visible table. An empty selection produces no demand edit.

\begin{promptbox}{Prompt A3: Target-selection instructions}
\begin{Verbatim}[breaklines=true,breakanywhere=true,breaksymbolleft={},breaksymbolright={}]
OD-list condition suffix:
For scale_od, inside the attribution object set the target_od_pairs key to the affected [origin, dest] pairs (two-element lists of the identifiers returned by tools). The target_zones key is descriptive only; it does not select OD rows.
Origin-selector condition suffix:
For scale_od in this condition, inside the attribution object set the target_origins key to an array of affected integer origin identifiers and set the target_od_pairs key to [] or omit it. The public executor expands each selected origin to every positive-demand OD row in visible_demand_table; select origins only, never destinations. The target_zones key remains descriptive metadata.
Shared output contract:
Return exactly one JSON object and nothing else (no Markdown fences, preamble, trailing prose, or second JSON object). The envelope must be {"action":"attribute","attribution":{...}}; the representation-specific field is nested inside attribution. All target_zones and target_sensor_ids are JSON arrays of integer identifiers, never prose; target_od_pairs is a JSON array of two-integer arrays (use [] when unused); road_class is a string or null; put explanations in rationale. For intervention_type none, all target arrays are [].
\end{Verbatim}
\end{promptbox}

\textbf{Generation and repetition.} K3, DeepSeek-Flash, and GLM-5.3 generate actions on four demand faults. Requests use JSON-object output and high reasoning where supported, with initial output limits of 49,152 tokens for K3 and GLM-5.3 and 65,536 for DeepSeek-Flash. Request deadlines are 3,600 seconds. Failed or truncated requests may be retried, with increased output allowances for truncation. Three repetitions of each model--fault--panel--interface combination produce 144 valid proposals. Each proposal uses only its assigned observations.

\textbf{Information separation.} Selection and evaluation use a fixed split of the 76 sensors into two panels of 38. Panel membership is fixed before proposal generation and shared across models and repetitions. Observations and diagnostic summaries supplied to the model use the selection panel and current demand table. Reversing the panels uses a separately generated pair of proposals.

\textbf{Candidate construction and scoring.} Each pair of proposals defines an OD-pair candidate and an origin-grouped candidate. Component crossing adds the two exchanged target--update combinations. Interpolation instead adds $(S_1,\bar u)$ and $(S_2,\bar u)$, where $\bar u=\sqrt{u_1u_2}$. Both expanded sets retain the originals and contain four candidates. Every candidate executes from the same pre-action state, using the same assignment solver and a $10^{-5}$ relative-gap criterion. Selection-panel loss determines the candidate before evaluation-panel scoring. The 72 directional evaluations yield 36 model--fault--repetition means by averaging the two directions.

\end{document}